\documentclass{OAGM}
\OAGMarXiv{1505.01065}
\usepackage{graphicx}
\usepackage{graphics}
\usepackage{amsmath}
\usepackage[numbers]{natbib}
\usepackage[vlined,commentsnumbered,boxed]{algorithm2e} 

\title{The Minimum Spanning Tree of Maximum Entropy}

\author{Samuel de Sousa and Walter G. Kropatsch\\
  Pattern Recognition and Image Processing (PRIP)\\Vienna University of
  Technology, Austria}

\begin{document}
\maketitle

\begin{abstract}
In computer vision, we have the problem of creating graphs out of unstructured
point-sets, i.e. the data graph. A common approach for this problem
consists of building a triangulation which might not always lead to the best 
solution. Small changes in the location of the 
points might generate graphs with unstable configurations and the topology of
the graph could change significantly. After building the data-graph, one
could apply Graph Matching techniques to register the original point-sets. 
In this paper, we propose a data graph technique based on the Minimum
Spanning Tree of Maximum Entropty (MSTME). We aim at a data graph construction 
which could be more stable than the Delaunay triangulation with respect to small
variations in the neighborhood of points. Our technique aims at creating data graphs which
could help the point-set registration process. We propose an algorithm with a single free parameter that
weighs the importance between the total weight cost and the entropy of the 
current spanning tree. We compare our algorithm on a number of different
databases with the Delaunay triangulation.
\end{abstract}

\section{Introduction}

The problem of point-set registration often involves the construction of the
so-called data graph~\cite{hancock, luo}, which is the graph created out of
unstructured point-sets.
One possibility for registering point-sets using graphs is to formulate the
problem as Graph Matching~\cite{vento}. We are interested in exploiting some
topological properties of data graphs which would alleviate the registration
procedure. For instance, we could design a data-graph which
would decrease the search space when matching points.

The Iterative Closest
Point (ICP) algorithm~\cite{icp} is a powerful solution for the
registration of point-sets. Such an algorithm would try to find for each
point in one set the closest point in the other set. Therefore, we could avoid
checking all points if we have some prior knowledge. For instance, if we could
state that the point we would like to match has a certain degree value, we could 
evaluate only the points which match this
criteria and greatly reduce the search space for the registration. This would be
true for a non regular data-graph~\footnote{A graph in which all nodes have
the same degree value.}, otherwise we would still need to search the whole
space of points. 

We would like to obtain a graph with 
high variability in the degree distribution.
This high variability can be encoded by the entropy of the degree distribution.
In~\cite{desousa}, we have performed the registration of point-sets via the
Min-Weight Max-Entropy problem in which the generated  edge-induced subgraph had
the maximum entropy. In this work, we constrain our edge-induced subgraph to be
a tree. Our problem is called the
Minimum Spanning Tree with Maximum Entropy (MSTME).
By minimizing the total edge weight, we can allow some robustness of the data graph with 
respect to small deviations of the points, by maximizing the entropy, we
increase the variability of the degree distribution which could later ease the 
registration process. We evaluate the current approach under different levels of noise 
and compare the stability of the data-graph with a Delaunay triangulation. 

The remainder of this paper is organized as follows: The related work is
presented in Section~\ref{sec:related}. Section~\ref{sec:objective} discusses
our objective function and explains how the entropy is measured in the graph.
Our optimization strategy is explained in details in Section~\ref{sec:optimization}.
Our experimental setup is disclosed in Section~\ref{sec:experiments}. Our
conclusions and future work are available in Section~\ref{sec:con}.

\section{Related Work}
\label{sec:related}

A data graph can be
obtained in a variety of ways, it is common to use the
Delaunay triangulation for the registration of point-sets~\cite{hancock,luo,
gpd}. The Delaunay triangulation is based on the condition that no other point should lie inside the circumcircle of
any triangle. It is the dual of the Voronoi diagram which partitions
the embedding space into regions closest to the point set. In this work, we are
working on suitable alternatives which would allow more stability during the
registration.

Our cost function is built upon a Minimum Spanning Tree formulation and entropy
maximization. In this sense, our work is closely related to \cite{amuller} as
the authors also formulate an optimization problem focusing on the entropy of a
Minimum Spanning Tree. However, they are dealing with a different problem by
proposing an entropy estimator for clustering. In our work, we are
interested on the data graph construction problem and our cost function differs from 
theirs since our entropy is calculated on the degree distribution of the generated MST while
they estimate the data entropy based on the length of the spanning tree. 
 
Our problem can also be seen as a bi-criteria optimization problem.
\citet{cmst} proposed a Lagrangian relaxation to solve the
constrained minimum spanning tree problem, which is also a bi-criteria
optimization proven to be NP-Hard by~\citet{aggarwal}.
Neighborhood search and adjacency search heuristics for the bicriterion minimum
spanning tree problem were proposed by~\citet{andersen}.

\section{Objective Function}
\label{sec:objective}

In order to measure the diversity found in the degree distribution,
we can calculate the Shannon entropy ($\mathrm{H}$) for a graph $G(V,E)$ as
follows:
\begin{equation}
 \mathrm{H}(G) =  -\sum_{v \in V_{\ne}} p(v) \log_2(p(v)).
\label{eq:entropy}
\end{equation}
\noindent where $p(v)$ is the probability of finding a node with a degree 
of $v$ among all distinct degree values ($V_{\ne}$ is the set of distinct degree
values of $V$).
The entropy measures the uncertainty associated with a random variable. As
defined in Eq.~\ref{eq:entropy}, a high entropy
$\mathrm{H}(G)$ of a graph $G(V,E)$ would indicate a high
``variability'' in the distribution of nodes $V$. The converse is also true, a
low entropy means low variability, as in a $k$-regular graph whose 
probability $p(k) = 1$ and $log(1) = 0$. 

Given a point-set $\mathcal{X}$, we calculate the distance from each point to
all the other points and store those
distances in $W$. This generates a complete graph
$K_{|\mathcal{X}|}(\mathcal{X},W)$. We seek to find a vector $U \in
\{0,1\}^{|W|}$ in such a way that when $U_i = 1$, we add $W_i$ to our graph. We can formulate our cost function simply as:

\begin{equation}
\begin{aligned}
& \underset{U}{\text{minimize}}
& & \displaystyle \sum_{i = 1}^{|W|} W_i U_i - \lambda (\mathrm{H}(G[U]))\\
& \text{subject to}
& & U \in \{0,1\}^{|W|},\\
& & & \textstyle\sum U = |W|-1,\\
& & & G[U]\text{ is connected}.\\
\end{aligned}
\label{eq:objective}
\end{equation}

\noindent where $G[U]$ is an edge-induced subgraph from vector $U$. $\lambda$ is
a parameter which weighs the importance between the two terms: minimizing the
total edge and maximizing the graph entropy. The first constraint states that we
have a discrete problem, i.e. we either add an edge ($U_i=1$) or we do not add it. The second constraint is a
necessary condition for our graph to be a tree, however, we need the third
constraint to force the graph to be connected.

\begin{figure}[t]
  \centering\includegraphics[width=12cm]{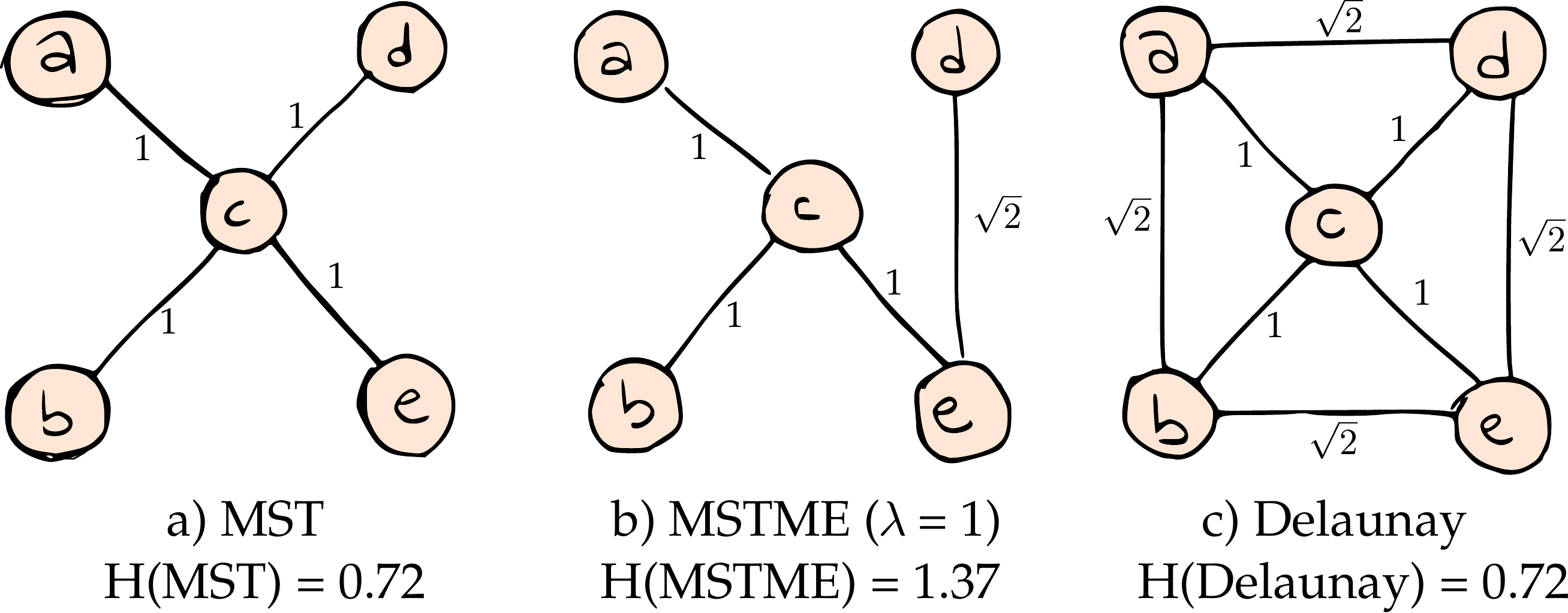}\\
  \caption{The first graph (a) is obtained with a Minimum Spanning Tree (MST)
  and has cost c(MST) = 4 and H(MST) = 0.72. Graph
  (b) is obtained using our approach of MST with Maximum Entropy (MSTME). Since
  H(MSTME) = 1.37, the cost c(MSTME, $\lambda=1$) = 3.04,
  hence, c(MST) < c(MSTME). The last graph (c) is a Delaunay triangulation and
  it has the same entropy as the MST.}
  \label{fig:mst}
\end{figure}

Figure~\ref{fig:mst} displays three different graphs of the same point-set
$\mathcal{X} = \left\{ a,b,c,d,e \right\} $. The first graph
(Figure~\ref{fig:mst}a) is simply a Minimum Spanning Tree (MST) whose cost is
$c(MST,\lambda=0) = 4$. It could be obtained by setting $\lambda=0$ and therefore the entropy part vanishes from Equation~\ref{eq:objective}. The
second graph (Figure~\ref{fig:mst}b) now considers $\lambda=1$, a higher entropy
would be able to decrease the cost of solution in comparison with the MST, the cost is
	c(MSTME,$\lambda=1$) = 3.04. Notice that although the edge weight cost is
	higher, the total cost was decreased due to the higher entropy.
	Finally, we show in contrast that calculating a Delaunay triangulation yields an entropy as high as the MST one.

\section{Optimization for the MSTME Problem}
\label{sec:optimization}

Our optimization is based on the following strategy: we pose the MSTME as 
a capacity problem with respect to the edge set. At each iteration, we increase 
the capacity of the edges, and we add the one whose cost is minimum based on
our cost function which evaluates the weight and
entropy of the current generated MST.

The input is a complete graph $K_{|\mathcal{X}|}(\mathcal{X},W)$ composed of
the point-set as the vertices and the edges are weighted according to the
Euclidean distance. For each edge $e$, we check if the 
subgraph $G[U]$ induced by our vector $U$ contains a cycle by the addition
of such an edge, in which case, we discard (line 8). Otherwise, we calculate
the new cost obtained by the weight of $e$ and we estimate the entropy of
the MST (line 12). In line 16, we remove this edge from our graph and continuing searching within the same capacity for an
edge which could yield a smaller cost. At the end, we add the edge with the
best cost an increase the capacity of our MST up to $|\mathcal{X}|-1$ (line 17).

\begin{algorithm}[H]
\LinesNumbered
\KwData{$K_{|\mathcal{X}|}(\mathcal{X},W)$, $\lambda$;}
\KwResult{$G[U]$;}
\DontPrintSemicolon
\Begin{
  $\text{U} \leftarrow \{\};$

  \tcp{We increase the capacity up to $|\mathcal{X}|-1$}
   \For{$i=0$ \KwTo $|\mathcal{X}|-1$}{
   
   	 $cost \leftarrow \infty;$
   	 
   	 $best \leftarrow 0;$
   	 
   	 \ForEach{$e \in E \setminus U $}{
   	 
   	 \uIf{$G[U \cup \{e\}]$ has cycle}{ 
   	 	\textbf{continue};
   	 	
   	 }\Else{ 
   	 	$U \leftarrow U \cup \{e\};$
   	 }
   	 
   	 $\text{ecost} = W_e - \lambda H(G[U])$  
   	 
   	 \If{ecost < cost} {
   	 	$cost \leftarrow ecost$;
   	 	
   	 	$best \leftarrow e$;
   	 }
   	 
   	 $U \leftarrow U - \{e\};$
   	 }
   	 
   	 $U \leftarrow U \cup \{best\};$
	}
	\textbf{return} $G[U]$;
} 
\caption{The Minimum Spanning Tree with Maximum Entropy.}
\end{algorithm}

\section{Experiments}
\label{sec:experiments}

\begin{figure}[htp]
\begin{center}  
\begin{tabular}{@{}c@{}c@{}c@{}c@{}}
  \includegraphics[width=3cm]{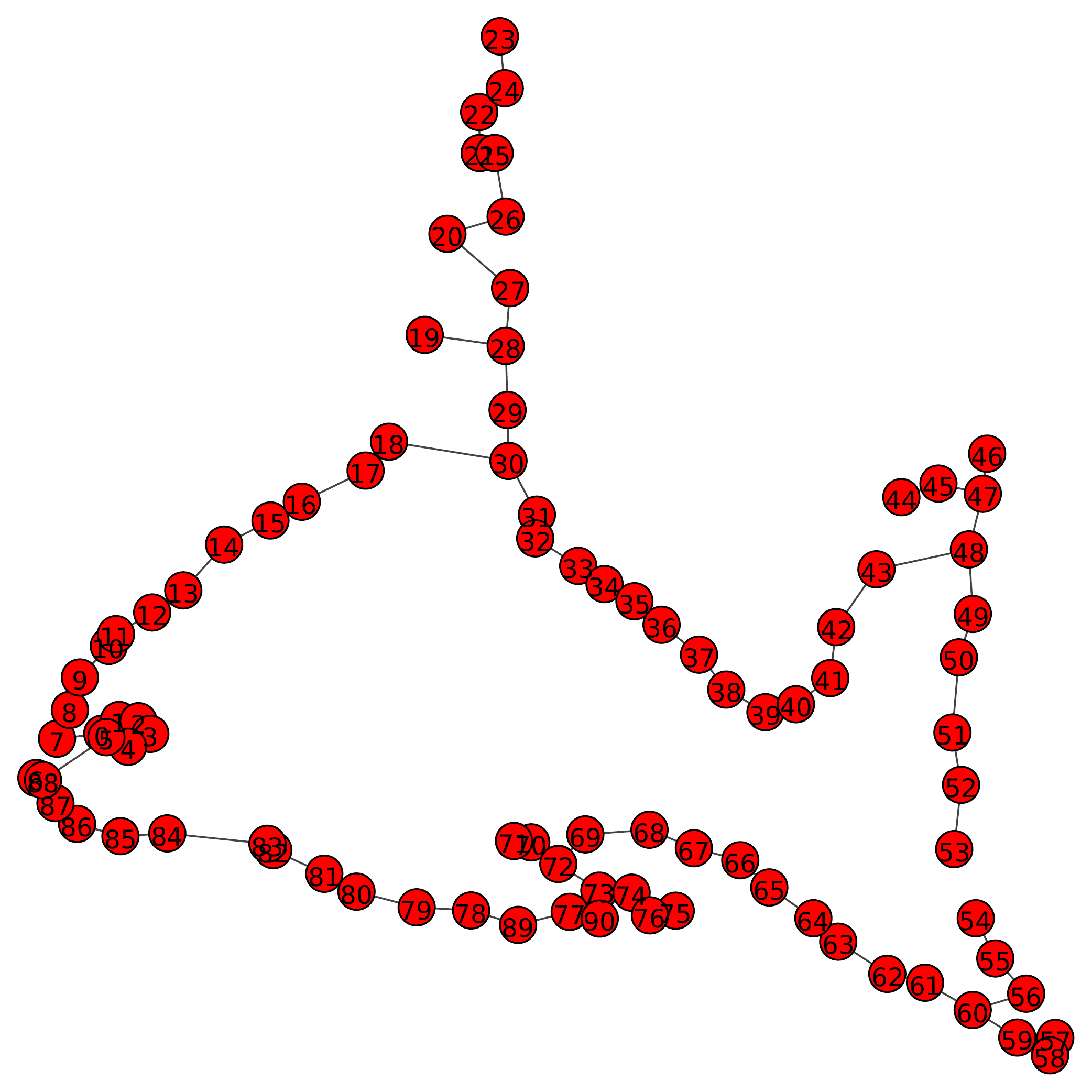}
  & 
  \reflectbox{\rotatebox[origin=c]{180}{\includegraphics[width=3cm]{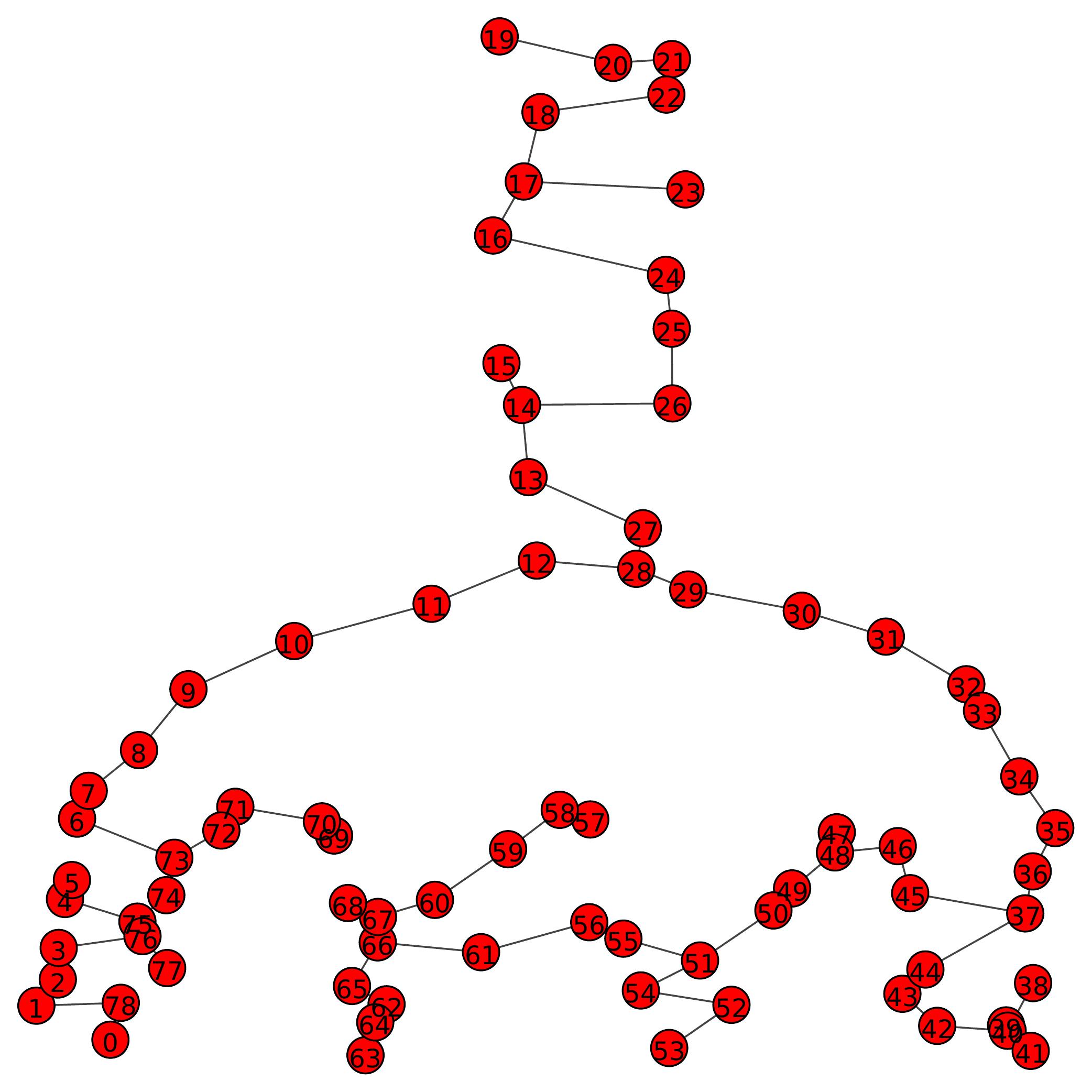}}} & 
  \includegraphics[width=3cm]{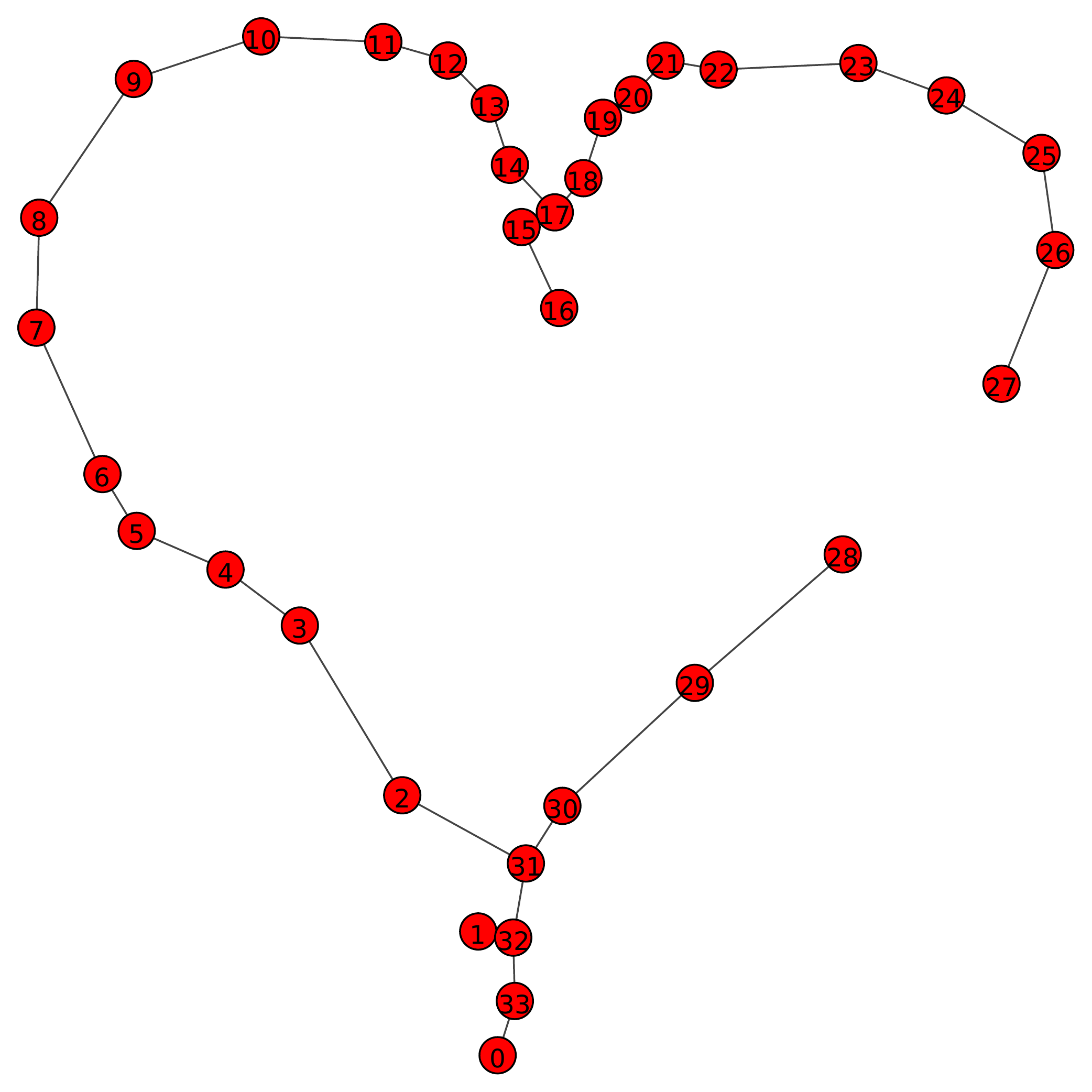}
  & 
  \reflectbox{\rotatebox[origin=c]{-90}{\includegraphics[width=3cm]{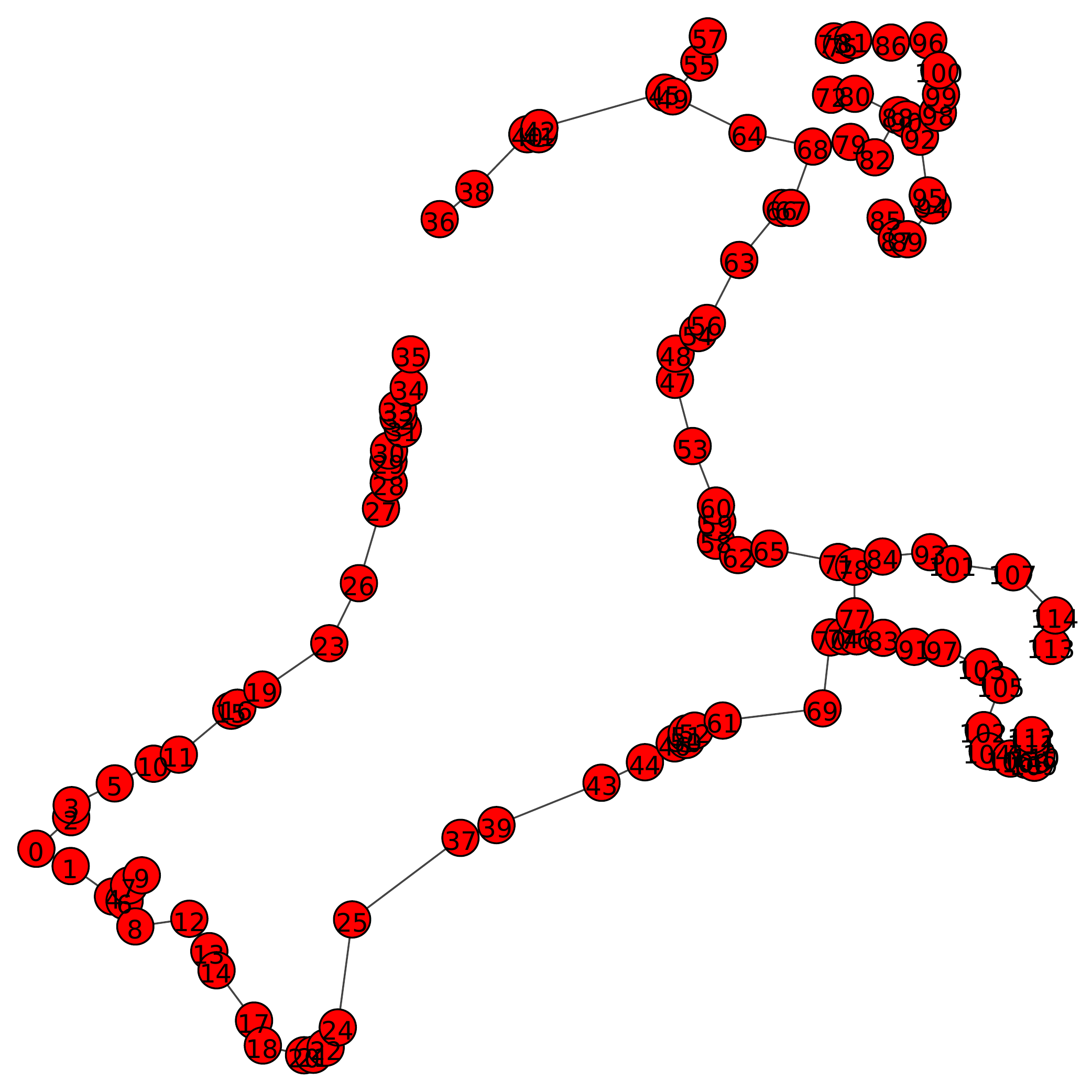}}}\\
\end{tabular}
  \caption{Data graphs connected using the MSTME ($\lambda=0.5$). }
  \label{fig:experiments}  
\end{center}
\end{figure}

Our experimental section focus on evaluating the stability of our algorithm
on point-sets sampled from the silhouette of objects.
Figure~\ref{fig:experiments} shows some point-sets connected using the MSTME. 
It is noticeable that the entropy takes over in thin areas, i.e. the degree
values of the points are higher in regions such as the top of the fish, the feet
of the dog, while other regions follow the silhouette of the object, since
  it is too costly to make an edge cross the shape of the object.

Our noise experiment is built upon the shortest pairwise
distance ($\epsilon$) in the point-set. Then, we perburb each point $x \in
\mathcal{X}$ by a length and an angle $(l,\alpha)$, where $\alpha=[0,360)$, and the radius
$l=[0,r.\epsilon]$ and $r$ varies from $[1,10]$, e.g. $5\epsilon$ means that
each point of the dataset was perturbed within a radius of 5 times the shortest pairwise
 distance in the graph. Certainly, the higher the noise,
the lower the stability of the algorithms will be.

\begin{figure}[htp]
\begin{center}  
\begin{tabular}{@{}c@{}c@{}c@{}c@{}} 
  \reflectbox{\rotatebox[origin=c]{-90}{\includegraphics[width=3cm]{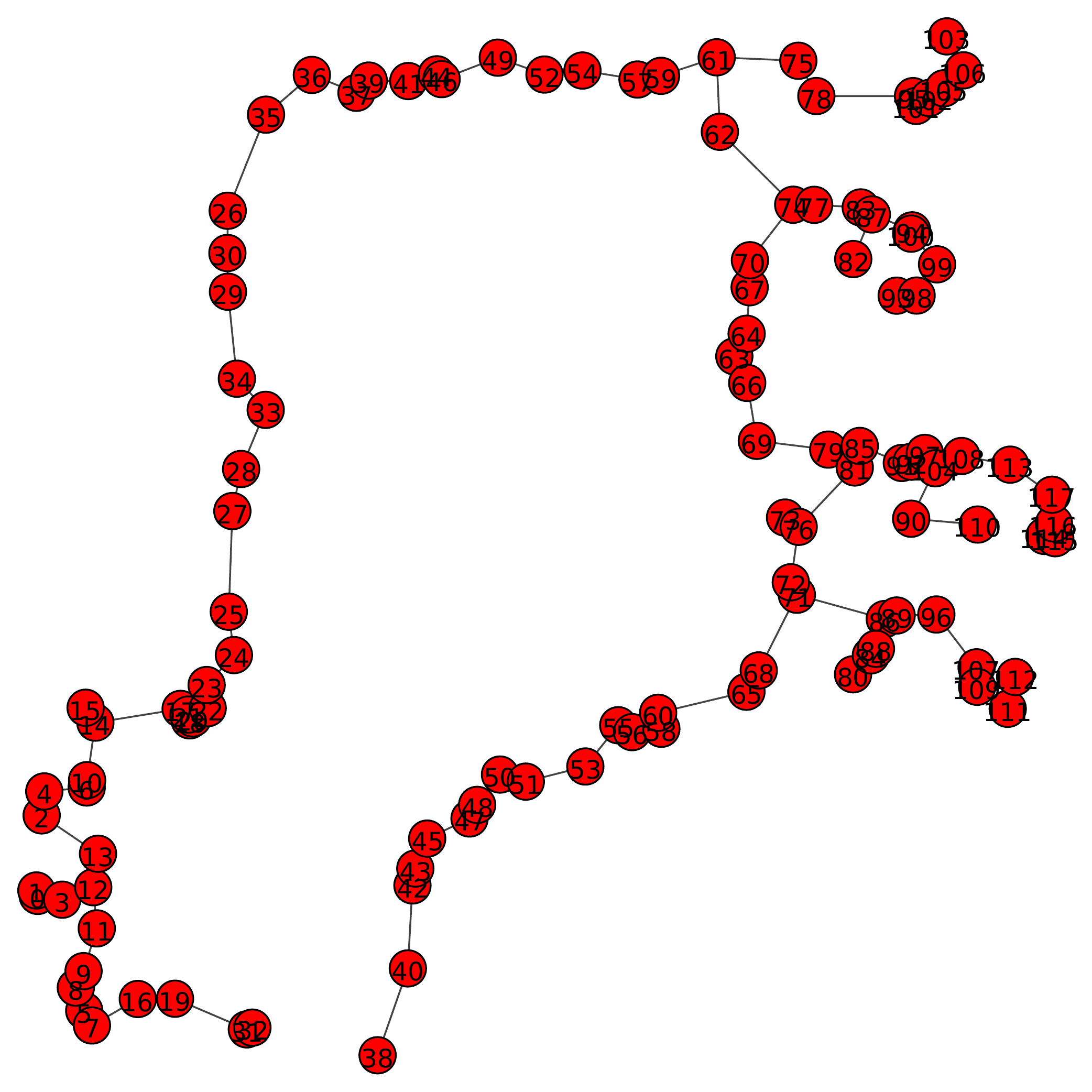}}}
  & 
  \reflectbox{\rotatebox[origin=c]{-90}{\includegraphics[width=3cm]{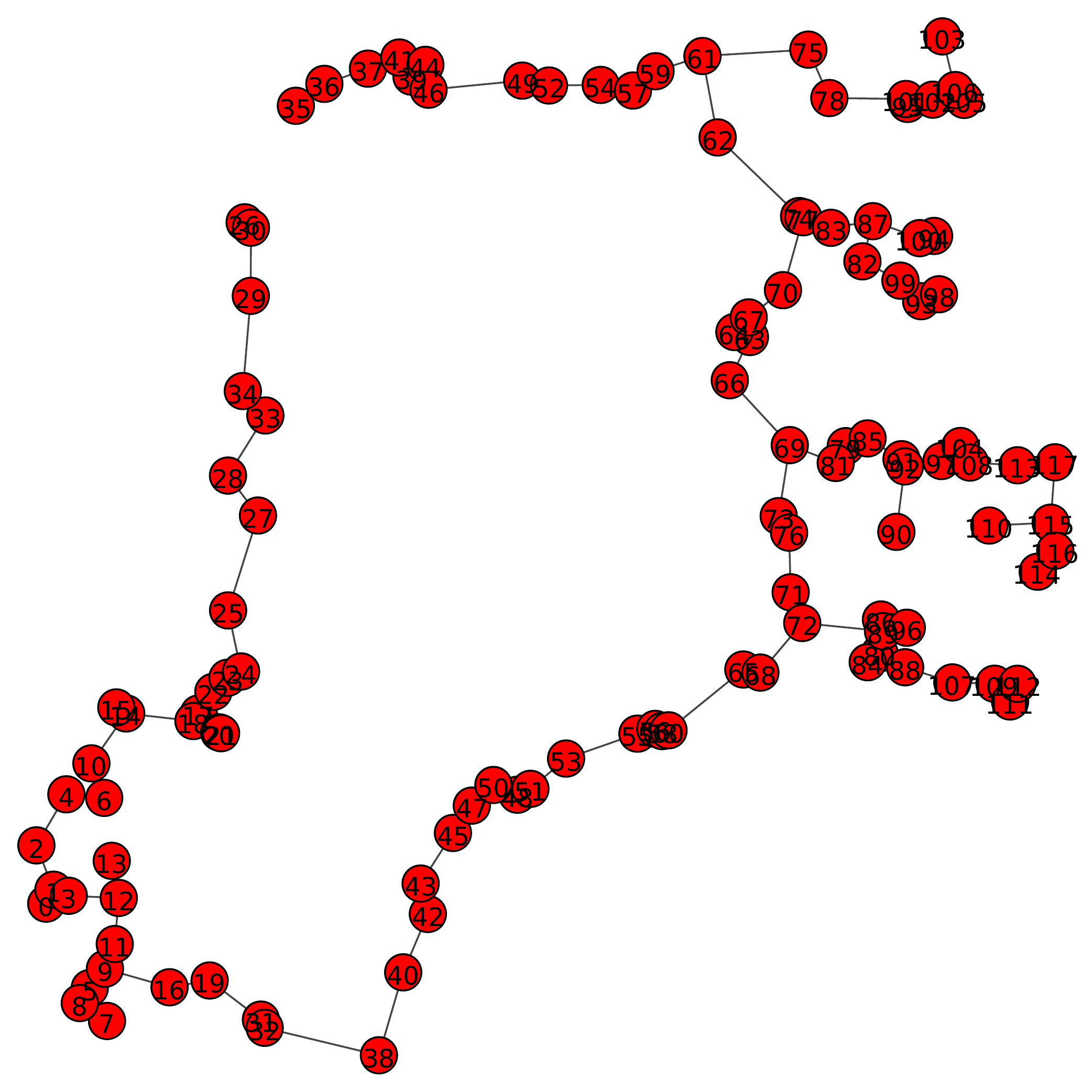}}}
  &
  \reflectbox{\rotatebox[origin=c]{-90}{\includegraphics[width=3cm]{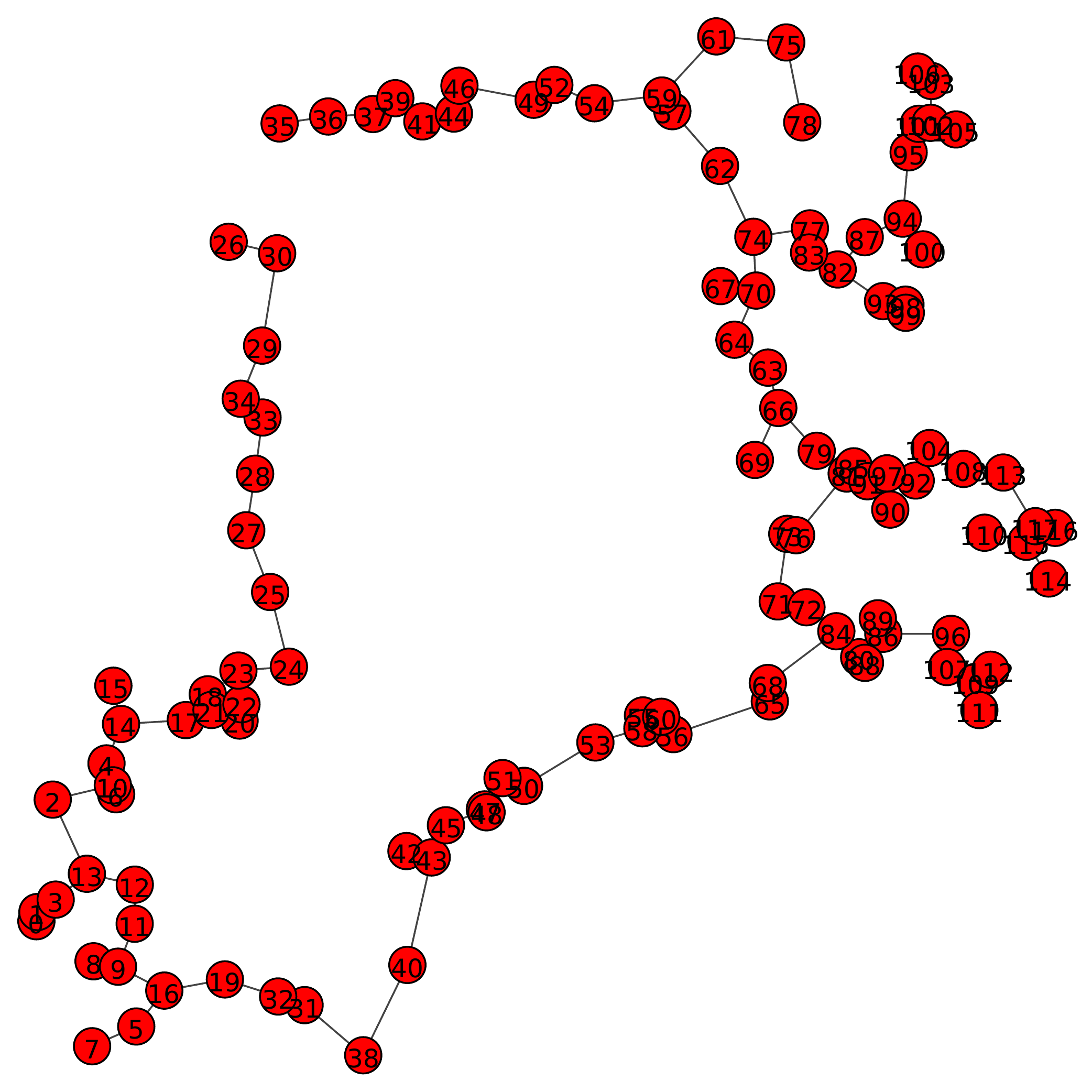}}}
  &
  \reflectbox{\rotatebox[origin=c]{-90}{\includegraphics[width=3cm]{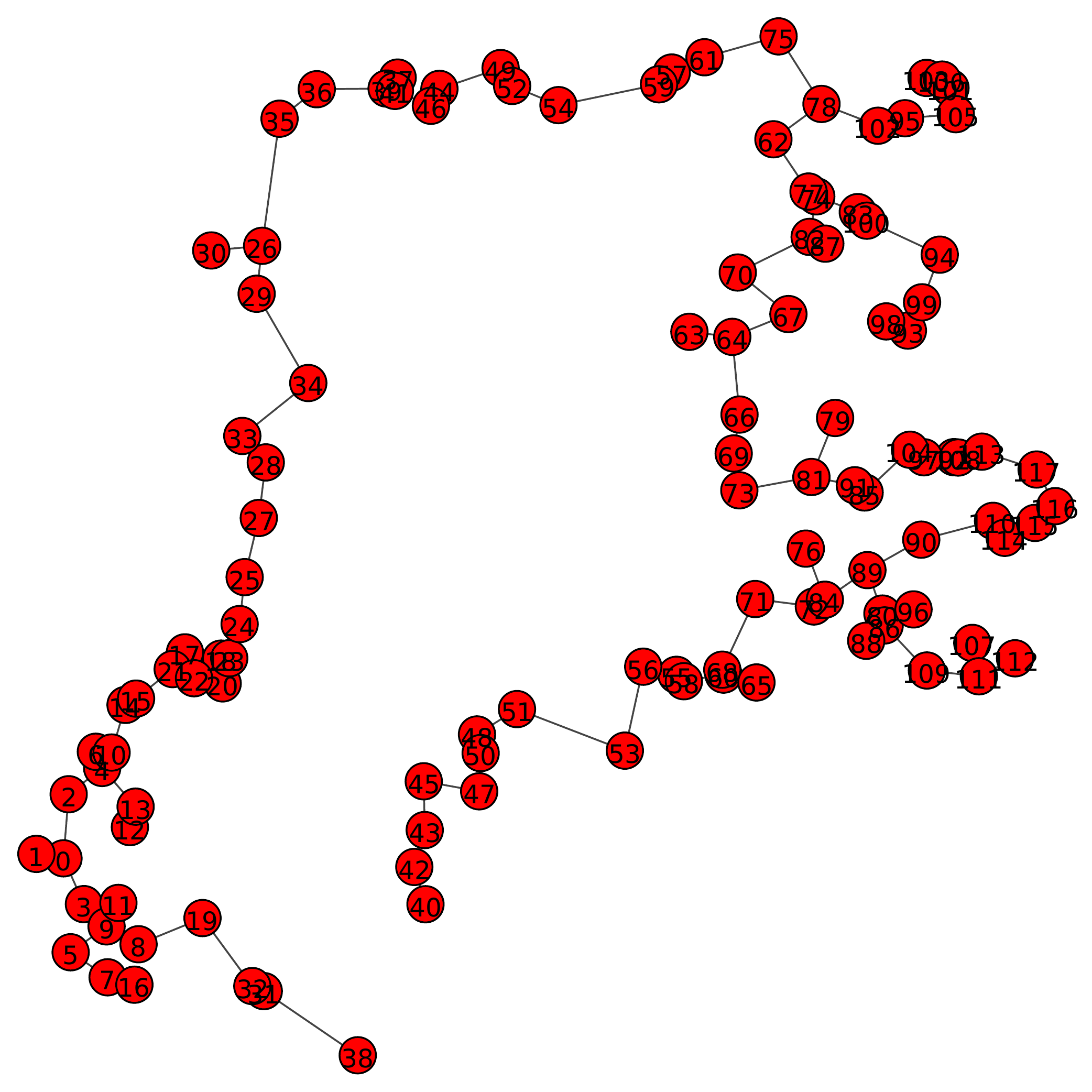}}}\\
  Noise up to $3\epsilon$ &Noise up to $5\epsilon$ &Noise up to $8\epsilon$
  &Noise up to $10\epsilon$ \\
\end{tabular}
  \caption{Noise added to the point-set.}
  \label{fig:noise}  
\end{center}
\end{figure}

Figure~\ref{fig:noise} shows the impact of noise in the point-sets varying
according to $\epsilon$ in a cow silhouette. It is not an easy task even
for humans to match such points (e.g. feet). It is worth mentioning that as the
algorithm searches for the optimal, it has to decide on which edges to take, therefore, a simple wrong edge will cause a much
higher impact on the results when compared with Delaunay triangulation, which
has many more edges.  

\begin{figure}[htp]   
\begin{center}
  \includegraphics[width=12cm]{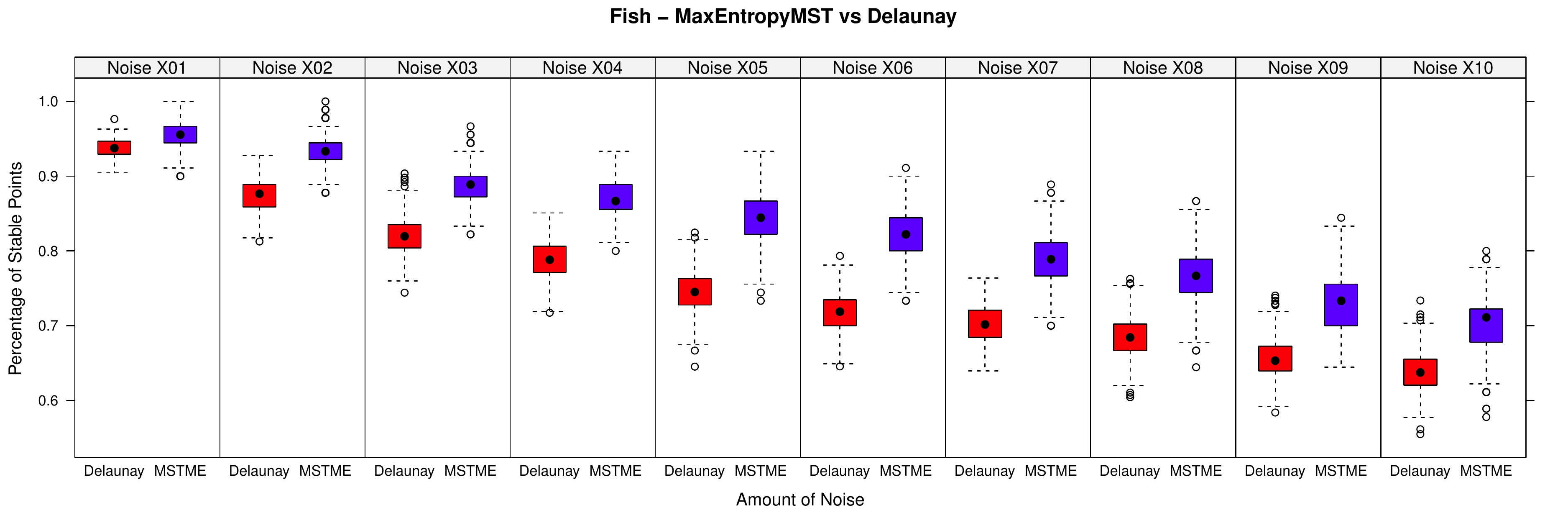}\\
  \includegraphics[width=12cm]{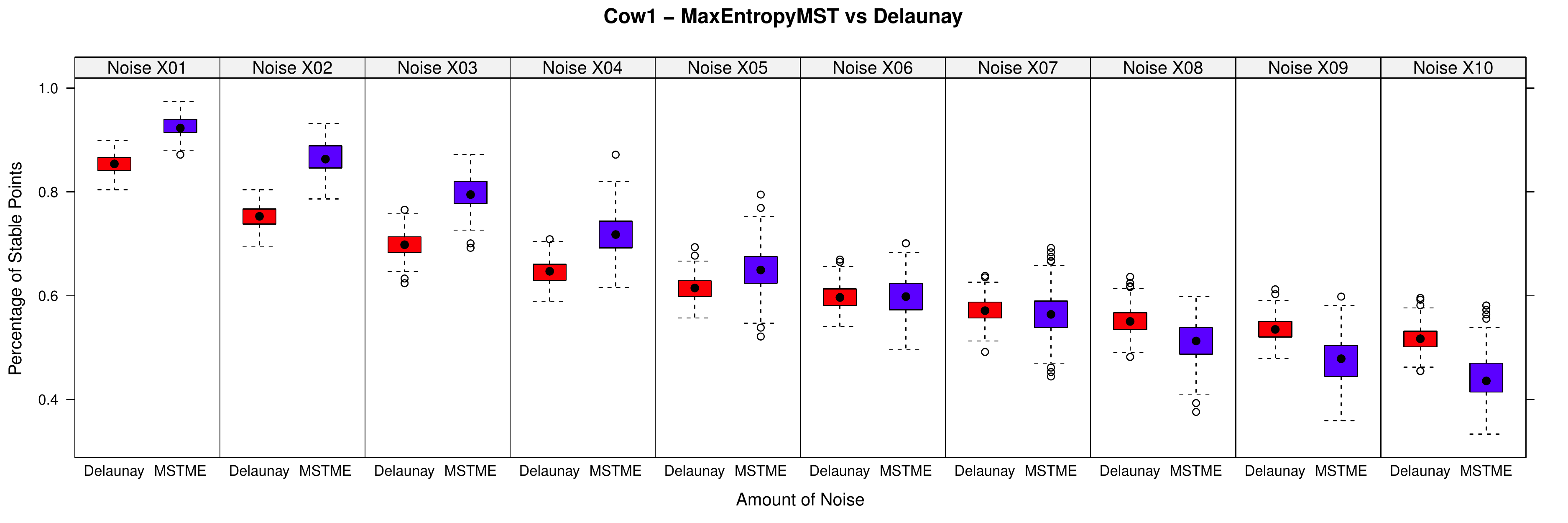}\\
  \caption[labelInTOC]{Boxplot comparison of the stability on the fish dataset.
  The noise consists of shifting each individual point by $\epsilon \times
  $level, varing from $1$ to $10$. For each noise level, we generated 30
  perturbed datasets and compute the number of stable edges accross all
  results.}
  \label{fig:boxplot_fish}
\end{center}
\end{figure}

As our algorithm outputs a
tree, it cannot contain cycles, therefore, the
silhouette will ``break'' in locations where the weights of the edges are high,
and the split point will vary according to the noise. Notice that the split
occurred in the face of the cow when noise level was $10\epsilon$ but in level $8\epsilon$ it happened at a
different location.

Figure~\ref{fig:boxplot_fish} shows the quantitative analysis of stability
between MSTME and Delaunay triangulation. The way we evaluate the stability is
by calculating the percentage of edges which remained stable in all graphs
within the same noise level. The first graph shows the results for the fish
dataset\footnote{The fish dataset was obtained from the Coherent Point
Drift~\cite{cpd}.}. The $y$-axis displays the percentage of stable edges,
e.g. 0.8 means that 80\% of edges were found across all data-graphs within the
same level. For this dataset, our algorithm for the MSTME was always more stable
than the Delaunay triangulation up to the highest noise level. The boxplot
displays several other types of information, such as the median, quartiles, etc.
The instability observed was found on the thin parts of the fish, such as the
dorsal and caudal fins. 
The second plot shows the results on the cow1 dataset\footnote{the cow1 dataset
was obtained from the kimia99 dataset~\cite{kimia}.} 

We observed that the MSTME was, in general, more stable than the Delaunay
algorithm. Small variations in the location of the nodes affected the
triangulation more than it affected the MSTME. As the triangulation generates
many more edges, a single wrong edge will have a high impact on the results of the
MSTME. Nevertheless, it was still able to obtain higher
stability than Delaunay up to a certain level of noise.

\section{Conclusions} 
\label{sec:con}

In this paper, we proposed a data-graph technique by searching for a tree whose
cost is minimum regarding the total edge weight and maximum regarding entropy of
the degree distribution.
We called this problem the Minimum Spanning Tree of Maximum Entropy (MSTME).
In point-sets sampled from the silhouette of an object, we observed that points 
lying in thin areas were often unstable when there was noise. After analyzing
and comparing the robustness of the generated graph, we notice that the
data-graph generated by the MSTME was more stable than the one by the Delaunay triangulation.

In the future, we would like to propose an approximation algorithm which could
guarantee bounds on the MSTME problem. Moreover, we will apply this data-graph
 in a framework similar to ICP and evaluate the accuracy of the results
obtained as well as comparing 
the results using other data-graph techniques.

\section*{Acknowledgments}

The first author of this paper is financially supported by the Vienna PhD School
of Informatics.

\bibliographystyle{plainnat}
\bibliography{refs}
\end{document}